\pdfoutput=1

\documentclass[11pt]{article}

\usepackage[final]{style/acl}

\usepackage{times}
\usepackage{latexsym}

\usepackage[T1]{fontenc}

\usepackage[utf8]{inputenc}

\usepackage{microtype}

\usepackage{inconsolata}

\usepackage{graphicx}

\usepackage{subcaption}
\usepackage{float}
\usepackage{hyperref}
\usepackage{url}
\usepackage{booktabs}
\usepackage{multirow}
\usepackage{amssymb}     %
\usepackage{amsmath}
\usepackage{xspace}
\usepackage[most]{tcolorbox}
\usepackage{pifont}

\title{Measuring Users' Mental Models of Speech Translation \\in Human-AI Collaboration}

\author{HyoJung Han $\quad$ Nishant Balepur  $\quad$ Jordan Boyd-Graber $\quad$ Marine Carpuat \\
University of Maryland, College Park, USA \\
\texttt{\{hjhan, nbalepur, jbg, marine\}@cs.umd.edu} %
}

\newif\ifcomment\commenttrue

\usepackage[a-1b]{pdfx}

\usepackage{framed}
\usepackage{mdwlist}
\usepackage{siunitx}
\usepackage{latexsym}
\usepackage{colortbl}
\usepackage{xcolor}
\usepackage{nicefrac}
\usepackage{booktabs}
\usepackage{fnpct}
\usepackage{amsfonts}
\usepackage[T1]{fontenc}
\usepackage{bold-extra}
\usepackage{amsmath}
\usepackage{amssymb}
\usepackage{bm}
\usepackage{graphicx}
\usepackage{mathtools}
\usepackage{microtype}
\usepackage{multirow}
\usepackage{multicol}
\usepackage{xpatch}
\usepackage{latexsym,comment}
\usepackage[normalem]{ulem}

\newcommand*{\missingreference}{{\Huge \colorbox{red}{?reference?}}}
\newcommand*{\missingcitation}{{\Huge \colorbox{red}{?citation?}}}

\makeatletter
\xpatchcmd{\@setref}{\bfseries}{\missingreference}{}{}
\def\@citex[#1]#2{\leavevmode
    \let\@citea\@empty
    \@cite{\@for\@citeb:=#2\do
        {\@citea\def\@citea{,\penalty\@m\ }%
            \edef\@citeb{\expandafter\@firstofone\@citeb\@empty}%
            \if@filesw\immediate\write\@auxout{\string\citation{\@citeb}}\fi
            \@ifundefined{b@\@citeb}{\hbox{\reset@font\missingcitation}%
                \G@refundefinedtrue
                \@latex@warning
                {Citation `\@citeb' on page \thepage \space undefined}}%
            {\@cite@ofmt{\csname b@\@citeb\endcsname}}}}{#1}}
\makeatother

\newcommand{\gem}[1]{\mbox{\textsc{gem}}}
\newcommand{\abr}[1]{\textsc{#1}}

\newcommand{\hidetext}[1]{}
\newcommand{\ignore}[1]{}

\ifcomment
    \newcommand{\pinaforecomment}[3]{\colorbox{#1}{\parbox{.8\linewidth}{#2: #3}}}

    \newcommand{\prtodo}[1]{\pinaforecomment{lightblue}{pr}{#1}}
    \newcommand{\prtodoi}[1]{\pinaforecomment{lightblue}{pr}{#1}}
\else
    \newcommand{\pinaforecomment}[3]{}
    \newcommand{\prtodo}[1]{}
    \newcommand{\prtodoi}[1]{}
\fi

\newcommand{\smallurl}[1]{ \begin{tiny}\url{#1}\end{tiny}}

\definecolor{lightblue}{HTML}{3cc7ea}
\definecolor{CUgold}{HTML}{CFB87C}
\definecolor{grey}{rgb}{0.95,0.95,0.95}
\definecolor{ceil}{rgb}{0.57, 0.63, 0.81}
\definecolor{UMDred}{HTML}{ed1c24}
\definecolor{UMDyellow}{HTML}{ffc20e}

\ifcomment
    \newcommand{\customcmt}[3]{\textcolor{#1}{[#2: #3]}}
\else
    \newcommand{\customcmt}[3]{}
\fi

\newcommand{\qa}{\textsc{qa}\xspace}
\newcommand{\mt}{\textsc{mt}\xspace}
\newcommand{\mm}{\textsc{mm}\xspace}
\newcommand{\mms}{\textsc{mm}s\xspace}
\newcommand{\ai}{\textsc{ai}\xspace}
\newcommand{\qe}{\textsc{qe}\xspace}
\newcommand{\io}{\textsc{i/o}\xspace}
\newcommand{\asr}{\textsc{asr}\xspace}
\newcommand{\xcomet}{\textsc{xcomet}\xspace}

\definecolor{SkyBlue}{rgb}{0.53, 0.81, 0.92}

\newtcolorbox[list inside=prompt,auto counter,number within=section]{prompt}[1][]{
    colbacktitle=black!60,
    fonttitle=\small,
    coltitle=white,
    fontupper=\footnotesize,
    boxsep=3pt,
    left=0pt,
    right=0pt,
    top=0pt,
    bottom=0pt,
    boxrule=1pt,
    #1
}

\begin{document}
\maketitle

\begin{abstract}

Millions of people use machine translation (\mt) tools daily, yet little is known about their perception of what systems can and cannot do.
This paper studies users' mental models of speech translation systems through a new framework based on cross-lingual question answering, where users either accept \mt output or request professional re-translation to answer questions based on the information presented in a foreign language.
By analyzing user behavior and accuracy trends across varying translation qualities, we examine to what extent they can predict where the system is likely to be wrong, and how this mental model evolves. 
Users develop stronger mental models with practice, especially when they have some knowledge of the source language, primarily by relying on surface-level error cues.
Moreover, providing speech transcriptions can help users develop better mental models.
Our results show the promise of cross-lingual question answering as a downstream task for studying \mt mental models, and advancing our understanding of human–\ai collaboration.

\end{abstract}

\section{Introduction}

\begin{figure}[t]
    \centering
    \includegraphics[width=0.42\textwidth]{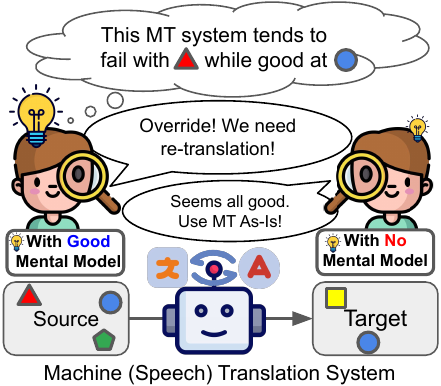} 
    \caption{Users' mental model, or their understanding of an \mt system's strengths and weaknesses, is essential for effectively integrating \mt system's output or appropriately intervening on it. Geometric shapes in the figure represent the features of input/output that the users can use in updating their mental model of \mt.
    }
    \label{fig:mm4mt_first}
\end{figure}

\begin{figure*}[t!]
    \centering
    \includegraphics[width=0.99\textwidth]{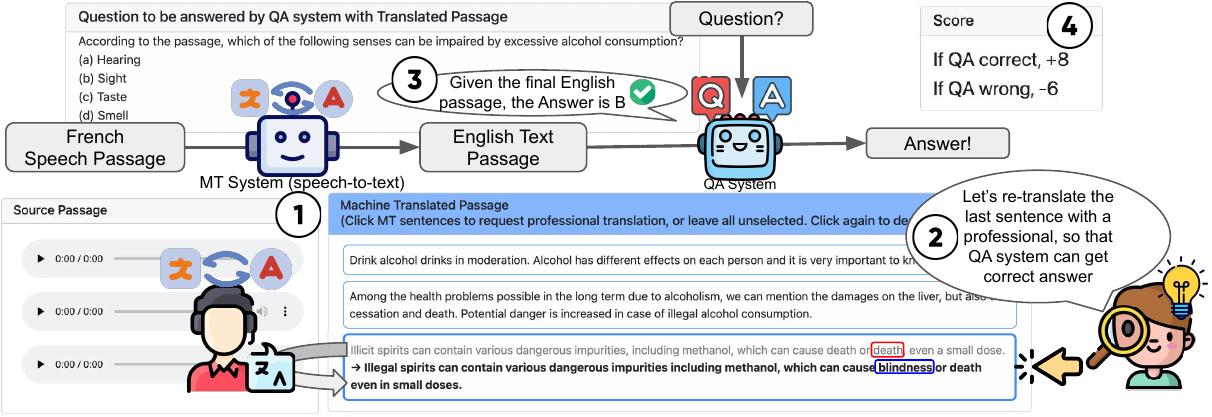} 
    \caption{
    Our proposed framework for measuring and updating users' mental models of \mt system.
    \raisebox{-0.25ex}{\large\ding{192}}: \mt system translates the source passage input.
    \raisebox{-0.25ex}{\large\ding{193}}: users decide which one among each segment to re-translate by a professional translator, and which one to use as-is, forming the final passage.
    \raisebox{-0.25ex}{\large\ding{194}}: the \qa system does the \qa consuming the passage.
    \raisebox{-0.25ex}{\large\ding{195}}: the user's reward is by the \qa correctness minus how much they re-translated.
    }
    \label{fig:mm4mt_interface_qa}
\end{figure*}

Millions of people use machine translation (\mt) tools daily, including in both casual \citep{calefato2016assessing, xiao2025sustaining} and high-stakes contexts where errors can have serious consequences \citep{liebling2020unmet, vieira2023ukmtusage}.
To use \mt effectively, users must understand inputs and scenarios where systems work reliably and where they fail, so they can make informed choices about e.g., when to trust outputs and when to seek human translation. 
This is particularly needed in \textit{speech} translation,  where audio inputs further amplify variability in output quality \citep{spechbach2019speech,speechqe} and reflect more realistic user scenarios. This understanding of a \mt system's strengths and limitations is a much needed form of \mt literacy~\citep{bokerciro2019mtliteracy, obrien2020mtliteracycogview}, and it forms the basis of users' \textbf{mental model}s~(\mms)---their internal understanding of how an \ai system works, when it succeeds or fails, and how to act on its outputs~\citep{norman1983some, hoffman-measures-for-xai}. Such models are critical for integrating \ai recommendations into decision-making~\citep{de-arteaga2020acasehumanintheloop, sieker-etal-2024-illusion}.

Despite its importance, the study of users' mental models in \mt remains limited. The \textsc{hci} literature provides a starting point for studying the role of \mm in human-\ai collaboration, but primarily focuses on classification and regression tasks, where models have simpler error boundaries~\citep{Bansal_Nushi_Kamar_Weld_Lasecki_Horvitz_2019, kelly-capturing-human-mm-of-ai}, and \ai predictions directly align with the users' decision needs~\citep{liu-etal-2024-beyond-human, vats2025surveyhumanaicollaborationlarge}.
As a result, existing methods do not easily port to \mt, where natural language outputs can be imperfect in many ways, and users must determine how to use them to make informed downstream decisions. 

This paper seeks to \textbf{understand and improve users' mental models of \mt} (Figure~\ref{fig:mm4mt_first}) by introducing a framework based on cross-lingual question answering (\qa), where users aim to maximize task performance by either accepting \mt output as-is or intervening with professional re-translation. This process naturally teaches them to recognize \mt's error patterns and refine their mental models (Figure~\ref{fig:mm4mt_interface_qa}). This downstream framing emphasizes the ``fitness for purpose'' of translations, and encourages users to do a cost-benefit analysis of the impact of potential errors on answers, rather than assessing \mt quality in isolation \citep{mehandru-etal-2023-physician, xiao2025sustaining}.
In this setup, users' efforts to maximize task performance naturally reveal how they form and refine their mental models over the course of the task. %

This framework lets us study human \mms for speech translation, and investigate how different factors impact their formation: users' language proficiency, input/output features they rely on, and types of interventions most effective for updating \mms. As we will see, users refine their \mms over time (\S\ref{sec:results_overall}): fluent and intermediate users had consistent improvements, but basic users struggled to update their models (\S\ref{sec:results_prof}). When examining error features, users were most responsive to output surface-level cues such as incompleteness, while topic-related errors remained the hardest to detect (\S\ref{sec:results_feat}). Finally, transcription explanation proved most effective by offering additional clues about error sources, while error span explanation boosted accuracy but encouraged over-reliance, limiting \mm development (\S\ref{sec:results_assist}).

Overall, this work introduces a concrete framework to measure users' mental models of \mt systems, and provides a foundation for exploring how to promote \mt literacy and appropriate \mm development in the wild, and for studying \mm for a broader range of generation tasks such as summarization, research, and dialogue.

\section{Mental Model for Machine Translation}\label{sec:method}
We first review how mental models have been defined and studied in prior work (\S\ref{sec:prelim_mental_model}), and then describe how we design our study to investigate mental models in the context of \mt (\S\ref{sec:methods_measure_mm}).

\subsection{Background and Prior Work}\label{sec:prelim_mental_model}

Conceptualizations of mental models have been introduced in contexts ranging from cognitive psychology \citep{norman1983some, klein2008macrocognition, iida2024integratingllmandmm} to explainable \ai \citep{miller2018explanationartificialintelligenceinsights, mueller2019explanationhumanaisystemsliterature, brachman2025buildingappmmaichat}. We adopt the definition of users' mental models of an \ai system as their understanding of its capabilities and limitations: its strengths, weaknesses, and the boundaries of its errors \citep{Bansal_Nushi_Kamar_Lasecki_Weld_Horvitz_2019, kelly-capturing-human-mm-of-ai, vats2025surveyhumanaicollaborationlarge}.

Previous studies place users in human–\ai collaborative tasks, where participants are rewarded for making decisions that reflect appropriate expectations of the system's successes and failures.
\citet{kelly-capturing-human-mm-of-ai} asked participants to predict an \ai agent's accuracy across rounds of 12 trivia questions, with rewards tied to how closely their estimates matched actual performance---thus capturing their mental model of the system.
In \citet{Bansal_Nushi_Kamar_Lasecki_Weld_Horvitz_2019, Bansal_Nushi_Kamar_Weld_Lasecki_Horvitz_2019}, participants were tasked with a binary decision of determining whether an object was defective, guided by recommendations from an \ai system. They could accept or override the recommendation, with monetary rewards reflecting correct choices and steep penalties for errors, simulating high-stakes decision-making.

However, they focus on a classification task in which the system directly provides a decision to the user, a setup that does not align with \mt.
We address this gap by proposing a new approach to measuring and developing users' mental models in \mt (Section~\ref{sec:methods_measure_mm}).

\subsection{Measuring Mental Models for MT Systems in Human+MT Context}\label{sec:methods_measure_mm}

A key question in our study is how to meaningfully assess users' mental models of \mt. There are several ways to do that. One straightforward suggestion is to present both the source and its translated version, then ask the user to evaluate how well the source has been rendered, similar to a quality estimation (\qe) task but with a human~\citep{Specia2010,callison-burch-etal-2012-findings}. The user's assessment can be compared to actual metric scores and rewarded if their predictions are close, as in \citet{kelly-capturing-human-mm-of-ai}.

However, this setup may not fully align with real-world usage, as what makes a translation ``good'' by standard metrics does not necessarily reflect what users find ``useful''.
In other words, standard metrics can overlook whether the \mt quality is ``good enough''. 
For instance, generic measures may flag case disagreement or other minor stylistic issues but overlook critical errors from a user's perspective~\citep{tomita-etal-1993-evaluation, krubinski-etal-2021-just, simqa}. 
If we want to measure usefulness, we need to assess ``fitness for purpose''---how well a translation supports a downstream task the user cares about---rather than simply how accurately it predicts a quality score \citep{hovy2002principles, bowker2010can,liu-etal-2024-evaluation}.
Thus, focusing on utility---while also prompting users to weigh the potential impact of translation errors on their decisions---may be more valuable than %
intrinsic definitions of translation quality, especially in \mt as an intermediate tool scenario~\citep{mehandru-etal-2023-physician, xiao2025sustaining}.

As an alternative to a \qe-based setup, \textbf{we propose to use a cross-lingual downstream task to measure the mental model of machine translation}, where users can naturally learn to recognize \mt's error patterns and refine their mental models. %
We choose reading comprehension question answering as a downstream task because it can measure translation quality by testing whether key information is preserved well enough to answer the question, thereby emphasizing quality as fitness for purpose \citep{Jones2005MeasuringTQ,moghe-etal-2023-extrinsic,sweta2024taclassessrole, ki2025isharetranslationevaluating}.
The user sees the non-English source information and its translated English version.
The goal is to complete a \qa task using the translated English passage as effectively as possible.
But the translation is not static: for each translation segment (here, individual sentences), users can either accept the \mt output as is or request a re-translation by a ``professional'' (i.e., get a gold reference translation).

To control for the possibility that users might answer questions using world knowledge rather than the translation, we rely on a fixed \qa system, which isolates the influence of their \mms of \mt.
The reward is determined by whether the \emph{\qa system} answers correctly given the final translated input.
To prevent users from getting professional translations for everything, the reward decreases every time they request a re-translation from a base reward. %

\section{Key Research Questions}\label{sec:method_rqs}

We explore three primary research questions: how language proficiency affects the evaluation and development of mental models (\S\ref{sec:method_langflumm}), which input/output features users rely on to develop and update their mental models (\S\ref{sec:method_feat}), and what types of explanation are most effective for updating a mental model (\S\ref{sec:method_information}).

\subsection{RQ1: How Does Language Proficiency Affect Development of Mental Models?}\label{sec:method_langflumm}

Prior work shows that users who are not bilingual struggle to assess \mt outputs \citep{yimin2025emnlpplanetworld} and that interventions based on quality estimation and backtranslation affect reliance inconsistently, depending on fluency \citep{mehandru-etal-2023-physician, zouhar-etal-2021-backtranslation}. In this work, we study 
how mental models develop as a function of source-language proficiency. %
This can help us understand user decisions at a finer-grained level and inform training strategies to promote more appropriate reliance in the future.

\subsection{RQ2: Which Input/Output Features Do Users Rely On to Update Their MM?}\label{sec:method_feat}
To understand which characteristics they usually rely on to detect errors and update their mental model, we categorize input and output features of naturally occurring speech translation errors.
We focus on four main categories: %
translation errors containing \textit{rare words or named entities}, inputs with \textit{phonetic ambiguities or noise}, \textit{domain-specific errors}, and outputs that exhibit \textit{incompleteness or unnaturalness}.\footnote{We identified these four major categories based on our preliminary error analysis and prior literature about \mt errors. More details in Section~\ref{sec:exp_cur}}

Source speech containing \textit{\textbf{rare words or named entities}} often leads to errors in translation systems, particularly when contextual cues are limited~\citep{gaido-etal-2021-moby}. \textit{\textbf{Phonetic ambiguities or noisy inputs}} are also a key weakness of speech translation systems \citep{anwar2023muavic,xlavsr,mirzayev2025pronounciationintransl}. As noted by \citet{irvine2013taclmterrordomain}, \textit{\textbf{domain-specific errors}} occur when the system struggles more in certain topics (e.g., sports) while translating more reliably in others (e.g., science). Finally, \textit{\textbf{incompleteness or unnaturalness}} in translation is often the easiest indicator of error for users to detect \citep{martindale-carpuat-2018-fluency}.

While we do not explicitly prime participants with these features during the task, we also gather their reflections in a post-survey on ``what characteristics they have learned to look for when identifying incorrect translations'' to cross-check whether the categories we observe in error analysis are also salient to participants in forming or updating their mental models of the system.

\subsection{RQ3: What Type of Explanation Is Most Effective for Updating Mental Models?}\label{sec:method_information}

Providing appropriate explanations can improve users' ability to understand and predict system behavior, as clear feedback can help them make sense of complex models \citep{rutjesXAIMM2019, xie-etal-2022-calibrating, boyd-graber-etal-2022-human}. Building on this, we investigate which types of translation explanations most effectively help users build and refine their \mms of an \mt system.

Among several types of explanations that a machine translation model can generate by itself, we choose \textbf{transcriptions} as internal explanations that can be directly produced by the speech translation model.
The transcription provides insights into how the model processes and represents the source inputs. Even when the transcription is not explicitly used to generate translation, it offers clues for errors stemming from wrong audio representation and its possible improvements in translation \citep{dou2025speechasrst}. 

Beyond explanations generated by the speech translation model itself, external models such as quality estimation (\qe) can also provide insights. These models output segment-level scores and highlight error spans, helping users spot unreliable translations. In our setup, we use \textbf{error spans}---highlighted segments of the translation that contain errors---as external explanations \citep{zouhar-etal-2021-backtranslation,briakou-etal-2023-explaining}, since an overall score alone is often not reliable or useful to support users' understanding and \mm development \citep{mehandru-etal-2023-physician, ki2025isharetranslationevaluating}.
We compare a baseline with no explanations against conditions that provide transcriptions and error spans as local explanations.

\section{User Study Design}\label{sec:exp}
This section outlines the user study setup to address our research questions. We describe how we curated the question set to present to participants for effectively evaluating and developing their mental models, considering translation error features~(\S\ref{sec:exp_cur}). We then detail the interface in our experiments (\S\ref{sec:exp_interface}). Finally, we describe the participants and 
their proficiency groups (\S\ref{sec:exp_qe}).

\subsection{Curation of \qa Set}
\label{sec:exp_cur}

We curated 16 reading comprehension sets from 2\textsc{m}-\textsc{belebele} \citep{2mbelebele}, where each passage consists of French audio sentences and is paired with an English multiple-choice question.
We begin with question sets that Mistral-7B\footnote{\href{https://huggingface.co/mistralai/Mistral-7B-Instruct-v0.3}{\texttt{mistralai/Mistral-7B-Instruct-v0.3}}}\citep[as \qa model]{mistral7b} answers correctly when provided with a gold English passage.
Each French speech sentence is translated into English text using Whisper \citep{radford2022whisper}, a widely adopted speech translation system. 
We then provide the translated passage along with the corresponding multiple-choice question to the \qa model to do the reading comprehension task and record the system's multiple-choice response.
We further apply category classification\footnote{\href{https://huggingface.co/WebOrganizer/TopicClassifier-NoURL}{\texttt{WebOrganizer/TopicClassifier-NoURL}}} \citep{wettig2025organize} to each passage and collect \xcomet scores \citep{guerreiro2023xcomet} based on gold English transcription and translation output.

Based on the set where the \qa model is correct with the gold English passage but fails with the translated English passage, we run a preliminary analysis of speech translation errors that cause the \qa system to fail and manually categorize them into groups. From the topic-wise analysis, translations in the Sports domain tend to score lower, while Science translations are higher. We thus narrow the scope to these two domains, treating Sports as more translation-challenging and Science as comparatively easier.
Finally, we identify four major categories, as described in Section~\ref{sec:method_feat}.
Among the questions where translation errors lead to incorrect \qa answers, we select 11 examples in which the feature categories are evenly represented, except for the topic, which is dominated by Sports. In contrast, we select five correctly translated questions, which are largely drawn from the Science domain.

\subsection{Experiment Design}\label{sec:exp_interface}

\paragraph{Interface.} 
We implement an interface to train and test users' mental models of \mt.
The interface presents segment-level pairs of French audio and machine-translated English text along with reading comprehension multiple-choice questions (Figure~\ref{fig:mm4mt_interface_qa}).
Users are instructed to select only translations that they believe require professional re-translation to ensure the \qa system will produce correct answers based on the translated input.
\paragraph{Rewards.}  
If the \qa system answers correctly with the final passage, the user's score increases by the calculated reward. 
However, if the user fails to identify the problematic sentence containing a critical error causing a \qa error (recall that the \qa system by construction is correct on gold question translations),  they \emph{lose} six points.
To discourage excessive re-translation requests, each request deducts points proportionally from the base reward of 12, depending on the number of sentences selected (e.g., selecting one of four sentences lowers the reward to nine, while selecting two lowers it to six); if all sentences are selected, the reward collapses to zero.
The \textit{maximum possible score} for each question is therefore achieved by re-translating only the minimal necessary segment (usually one).

\paragraph{Explanation.} We also experiment with additional forms of explanation---transcriptions and error spans (Section~\ref{sec:method_information})---to examine what types of explanation are most effective for updating a mental model. 
\xcomet extracts the error spans based on the gold English transcription and the speech-translated text, and transcriptions are generated by the same speech translation model.
Users are automatically assigned to three different experimental conditions: Default (no explanation), Transcription, and Error span.
In the transcription condition, the text is displayed beneath the audio, while in the error-span condition, users see highlighted spans indicating the detected errors in the translated text.

\begin{table}[t]
\centering
\fontsize{9}{13}\selectfont
\begin{tabular}{ccccc}
\toprule
Assist ($\downarrow$)       & Basic & Intermediate & Fluent  & All \\
\midrule
Default       & 5     & 5            & 5      & 15   \\
Transcription & 4     & 6            & 5      & 15   \\
Error Span     & 4     & 7            & 4      & 15   \\
All           & 13    & 18           & 14      & 45   \\
\bottomrule
\end{tabular}
\caption{
Distribution of participants by French (source language) proficiency and the assistance experimental group with kinds of explanation. Most participants are proficient English (target language) speakers.
}
\label{tab:participants_dist}
\end{table}

\begin{figure*}[t]
\centering
\begin{subfigure}[t]{0.320\textwidth}
\includegraphics[width=\linewidth]{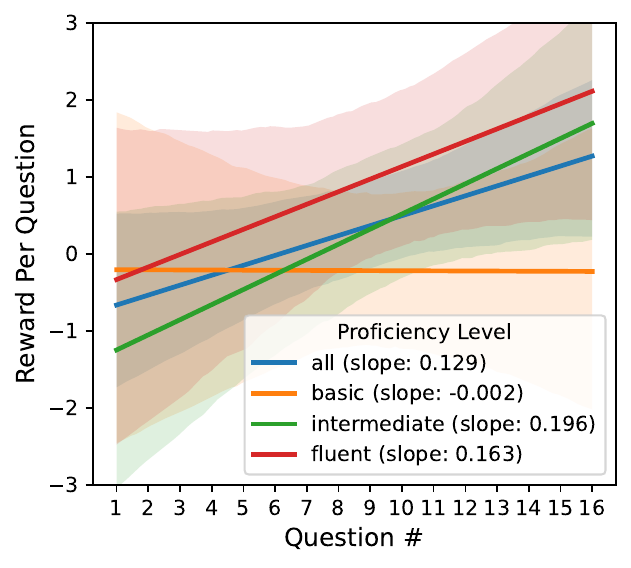}
\caption{Reward Trend ($\uparrow$)}
\label{fig:overall_scr_trend}
\end{subfigure}
\hfill
\begin{subfigure}[t]{0.33\textwidth}
\includegraphics[width=\linewidth]{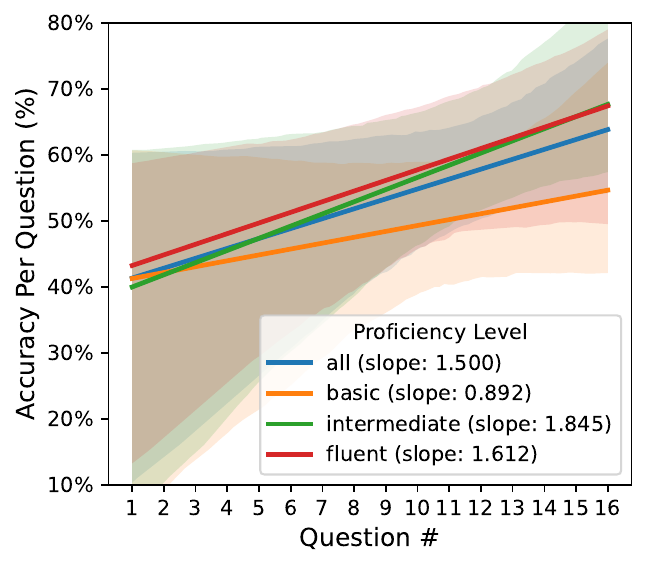}
\caption{\qa Accuracy Trend ($\uparrow$)}
\label{fig:overall_acc_trend}
\end{subfigure}
\hfill
\begin{subfigure}[t]{0.320\textwidth}
\includegraphics[width=\linewidth]{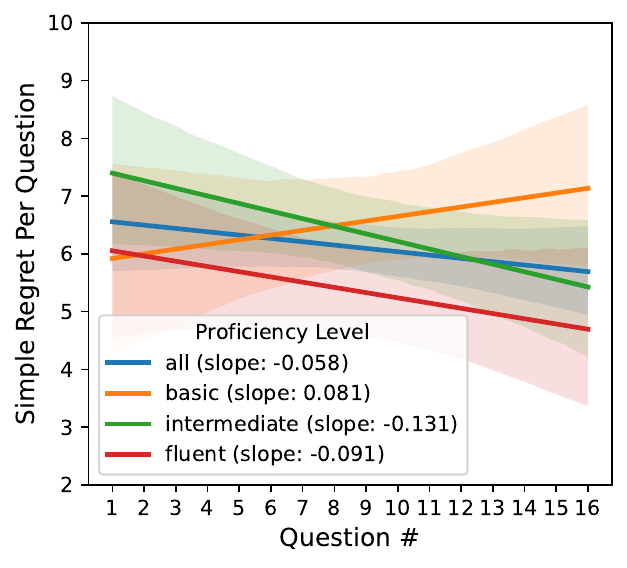}
\caption{Simple Regret Trend ($\downarrow$)}
\label{fig:overall_simreg_trend}
\end{subfigure}
\caption{
As participants progress, reward and accuracy increase (left, middle). Simple regret (maximum possible reward – actual reward) decreases as users become more precise in selecting only the problematic sentences (right).
The slope values reflect the average change with each additional question.
Participants progressively learn about the \mt system over the course of the task, developing better mental models.
}

\label{fig:overall_trends}
\end{figure*}

\subsection{Defining Source Language Proficiency}
\label{sec:exp_qe}

We ran user studies with crowd workers on Prolific\footnote{\url{https://www.prolific.com/}}, representing a diverse range of source-language fluency. The post-survey asks participants about their French proficiency.
We also include four quality estimation (\qe) tasks, presented before every four \qa tasks, where participants rate the adequacy of the translation on a 1--6 scale (Figure~\ref{fig:mm4mt_interface_qe}). We compare their \qe ratings with the converted \xcomet scores to assess user performance.
Based on their self-reported proficiency and \qe performance, we divide participants into three groups: basic, intermediate, and fluent (Table~\ref{tab:participants_dist}; \qe setup and proficiency grouping detailed in Appendix~\ref{sec:append_add_qe_prof}).

\section{Results: Participants Build MMs of MT}\label{sec:results}
This section presents our main findings on how users form and refine their mental models of the \mt system.
Users build better mental models as they gain experience with the system and the task~(\S\ref{sec:results_overall}). 
Next, higher language proficiency leads to better mental models (\S\ref{sec:results_prof}).
We then investigate which input/output features users rely on when developing and updating their mental models (\S\ref{sec:results_feat}). 
Finally, we assess that transcription explanation is more effective in supporting mental model updates for speech translation than error span explanation (\S\ref{sec:results_assist}). 
Additionally, we present key phrases extracted from users' reflection notes (Appendix~\ref{sec:results_usernote}).

\subsection{Participants Gradually Update their MMs}
\label{sec:results_overall}

Figure~\ref{fig:overall_trends} shows overall trends in three key metrics across all participants: reward, question accuracy, and simple regret (defined as the difference between the maximum possible reward and the actual reward; lower is better).
Figure~\ref{fig:overall_scr_trend} and \ref{fig:overall_acc_trend} present regression plots of reward and accuracy per question (higher is better), while Figure~\ref{fig:overall_simreg_trend} shows a regression plot of \textbf{simple regret}. The slope value in the legend of each plot reflects the average change with each additional question.
Especially, a negative slope in simple regret indicates that the user is approaching the optimal score by requesting re-translations only when necessary and avoiding redundant or incorrect selections, signaling that they are refining their mental models. In contrast, a positive slope suggests limited or no learning.

Overall positive slopes in reward and accuracy and negative slopes in simple regret suggest that \textbf{\textit{users gradually refine their mental models}}, enabling them to better identify \mt errors that cause the \qa system to fail.

\begin{figure*}[t]
\centering
\begin{subfigure}[t]{0.49\textwidth}
    \includegraphics[width=\linewidth]{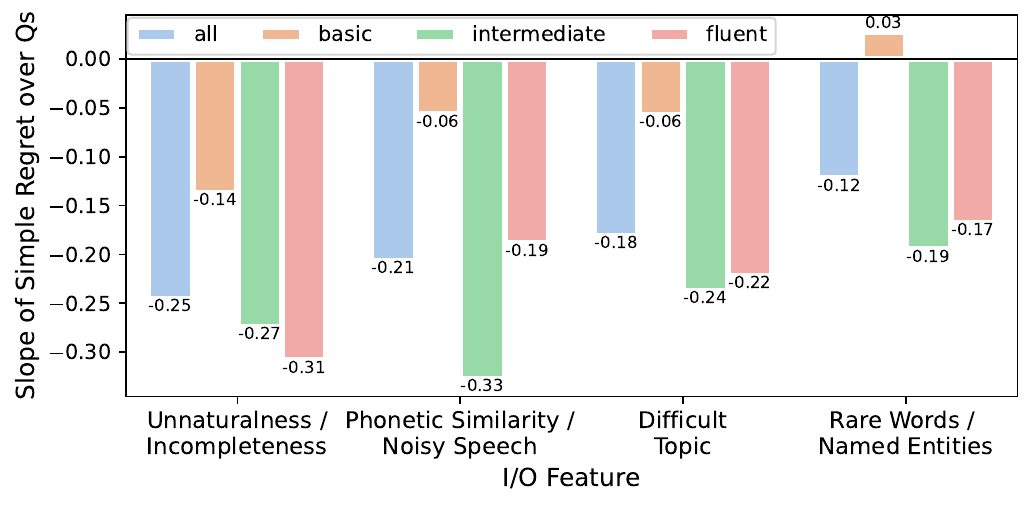}
    \caption{Slope of simple regret across question order.} 
    \label{fig:cond4_simreg_slope}
\end{subfigure}
\begin{subfigure}[t]{0.5\textwidth}
    \includegraphics[width=\linewidth]{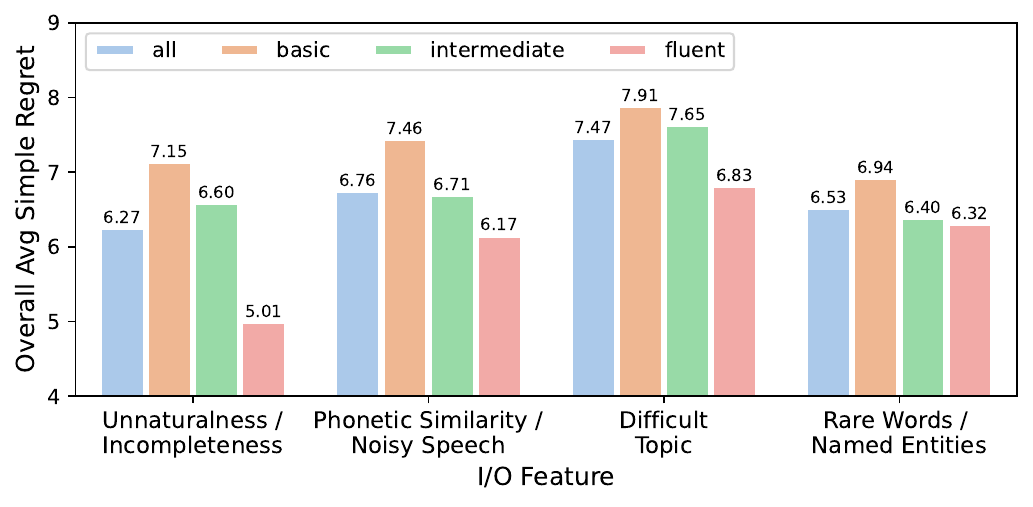}
    \caption{Overall average simple regret for each feature type.}
    \label{fig:cond4_simreg_absmean}
\end{subfigure}
\caption{
Simple regret (maximum possible reward – actual reward) trends within the questions associated with each input/output feature of \mt errors.
The slope value (left) reflects the average change with each additional question.
The \textit{incompleteness/unnaturalness} and \textit{phonetic similarity or noisy speech} with the steeper slope values are salient features of \mt errors that users readily pick up on to update their mental models, whereas \textit{topic-related errors} are the hardest for users to detect and leverage in this process.
}
\label{fig:cond4_simreg}
\end{figure*}

\subsection{A Little Bilingualism Goes a Long Way}\label{sec:results_prof}
While the overall trends suggest that users refine their mental models over time, Figure~\ref{fig:overall_trends} also highlights differences across proficiency groups. %

For reward and \qa accuracy, fluent and intermediate users both show stronger positive slopes, while the basic group shows a flatter (Figure~\ref{fig:overall_acc_trend}) or even negative (Figure~\ref{fig:overall_scr_trend}) slope.
In simple regret (Figure~\ref{fig:overall_simreg_trend}), the basic group shows an upward trend, which reflects that they struggled to properly develop a mental model of the \mt system; instead, they tend to select more sentences to avoid penalties from the incorrect \qa outcome.
Overall, the intermediate fluency group shows even steeper slopes compared to fluent users. One possible explanation is that fluent participants do well from the outset, even without fully developing their mental models, whereas intermediate users initially struggle but quickly learn to recognize the traits where the \mt system fails and improve more rapidly.
In summary, \textbf{\textit{higher source-language proficiency leads to better mental models}} of the translation system.

\subsection{Feature-Specific Trends in MM Updates}\label{sec:results_feat}
We group the question set according to the four input/output (\io) features of \mt errors categorized in Section~\ref{sec:method_feat}, and analyze the trends of simple regret within those groups to identify which features are most salient for users when updating their mental models (Figure~\ref{fig:cond4_simreg}).

We analyze both the slopes of simple regret across corresponding questions' order (Figure~\ref{fig:cond4_simreg_slope}) and the overall average simple regret (Figure~\ref{fig:cond4_simreg_absmean}). \footnote{The slope, as described in Section~\ref{sec:results_overall}, reflects the average change in performance per question, with negative values indicating a desirable trend toward optimal rewards.}
Across all users, \textit{incompleteness/unnaturalness} and \textit{phonetic similarity or noisy speech} show the steepest negative slopes, meaning that these features are easy for users to pick up and use when updating their mental models of \mt, allowing them to respond more effectively when encountering similar cases in the future.
In particular, the salience of \textit{incompleteness/unnaturalness} aligns with \citet{martindale-carpuat-2018-fluency}, which shows users are particularly reactive to disfluent translations.

\textit{Rare words or named entities} show the flattest slopes; however, their mean value (right-most bars in Figure~\ref{fig:cond4_simreg_absmean}) is already the lowest for basic and intermediate users, suggesting that users relied on this feature from the beginning.

The topic-related feature shows the second-flattest slopes and the highest overall average regret. We conjecture that this is because, even though participants are primed that the topic will be either sports or science, it is difficult for them to recognize the specific topic of a passage during the experiment unless it is explicitly indicated. As a result, topic-related errors are the hardest feature for users to detect and leverage when updating their mental models, while \textbf{\textit{incompleteness/unnaturalness} and \textit{phonetic noise} are the most salient to users}.

\subsection{Useful Explanation for Better \mm}\label{sec:results_assist}
Table~\ref{tab:score_acc_prof_assist} presents the results of different types of explanation---default (no additional support), transcription, and error span---on participants' final accumulated scores and overall \qa accuracy.\footnote{The plots with statistical significance analysis are in Appendix~\ref{sec:append_add_plots}, supporting meaningful difference between conditions and proficiency.}

\begin{table}[t]
\centering
\fontsize{9}{13}\selectfont
\begin{tabular}{ccccc}
\toprule
             \multicolumn{5}{c}{Final Score}       \\
Condition ($\downarrow$) & Basic & Intermediate & Fluent & All   \\
\midrule
Default       & 39.60        & 30.00  & 71.00 & 46.87 \\
Transcription & 34.00        & 56.67  & 54.20 & 49.80 \\
Error Span     & 41.75        & 34.14  & 43.00 & 38.53 \\
All           & 38.54        & 40.50  & 57.00 & 45.07 \\
\toprule
              \multicolumn{5}{c}{Overall Accuracy (\%)}  \\
Condition ($\downarrow$) & Basic & Intermediate & Fluent & All   \\
\midrule
Default       & 63.75 & 60.00 & 70.00 & 64.58 \\
Transcription & 60.94 & 71.02 & 70.00 & 67.99 \\
Error Span     & 68.75 & 71.43 & 68.75 & 70.00 \\
All           & 64.42 & 68.12 & 69.64 & 67.52 \\
\bottomrule
\end{tabular}
\caption{
Comparison of final scores and overall accuracy across three explanation conditions: default, transcription, and error span. Transcription provides the greatest benefit for intermediate users, while error spans increase accuracy but reduce rewards due to over-selection, limiting effective mental model development.
}
\label{tab:score_acc_prof_assist}
\end{table}

For final scores (upper), \asr transcription yielded the highest average across all users (49.80), while also improving overall accuracy (from 64.58\% to 67.99\%). This suggests that presenting the transcription alongside the translation provides users with additional clues about how translation errors may stem from incorrect audio representations, which in turn helps them better understand the speech translation system's behavior and develop more refined mental models.
The improvements are most prominent for intermediate French proficiency users, while performance for fluent users even decreases. One possible explanation is that achieving the highest level of performance requires careful listening to the audio, but the presence of \asr transcriptions may reduce the tendency to play and attend to the audio. As a result, fluent users might rely too heavily on the transcription, which limits their performance gains.

Error span explanation, on the other hand, yields the highest accuracy (70\%) but the lowest final score (38.53). This pattern arises because users often select any highlighted span, which increases correctness but also leads to frequent re-translation requests and consequently lower rewards, as similarly noted by \citet{sarti2025qe4pe}. While error spans help users achieve higher accuracy in this setup, they do not necessarily support the development of a robust mental model of the \mt system.

Overall, \textbf{\textit{transcription explanation is more effective, especially for intermediate proficiency users}}, while error span is less suitable for supporting the development of well-formed \mms.

\section{Analysis on User's Reflection}\label{sec:results_usernote}
\begin{figure}[t]
\centering
\includegraphics[width=0.48\textwidth]{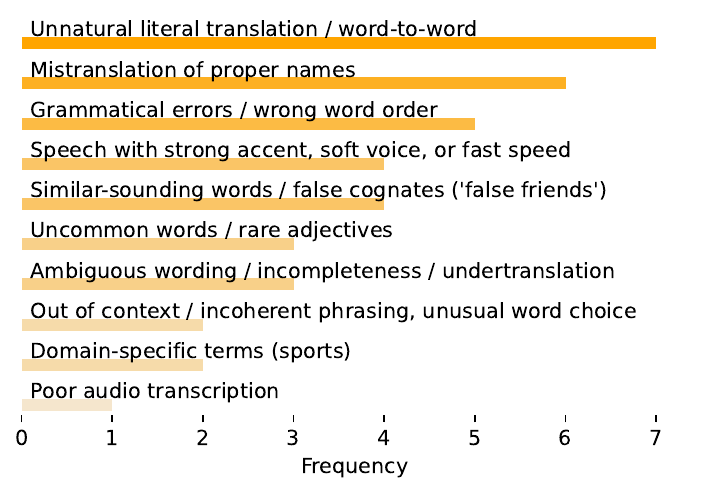} 
\caption{
Key phrases from participants' reflections on the speech translation system's error patterns and their identifying cues for updating their \mms. These align with our predefined input/output feature categories.
}
\label{fig:reflection_note}
\end{figure}
At the end of the survey, participants were asked to reflect on the types of errors the \mt system is likely to make and the characteristics they learned to look for when identifying incorrect translations. We manually analyzed the key phrases in these reflection notes and grouped them into categories, shown in Figure~\ref{fig:reflection_note}. The most frequently mentioned issue was unnatural literal translation, aligning with our earlier analysis of the most salient features (\S\ref{sec:results_feat}). Overall, we find that the traits users identify when updating their mental models align closely with our predefined categories of \textsc{i}/\textsc{o} features.

\section{Related Work}\label{sec:related_work}

\paragraph{Mental Model.}
A mental model can be understood as a general cognitive construct that people form through their interactions with the world, including others, themselves, the tasks they do, and the topics they learn. These models provide both predictive and explanatory power for making sense of such interactions \citep{norman1983some, kumar2023differentiating}.
In explainable \ai, a mental model refers to how users understand both the functioning of the \ai system and the conditions under which it is applied \citep{miller2018explanationartificialintelligenceinsights, mueller2019explanationhumanaisystemsliterature}. %

Mental models are closely linked to trust and transparency in \ai systems, highlighting the importance of users' understanding in anticipating system failures \citep{vats2025surveyhumanaicollaborationlarge}. Strengthening this mutual understanding between humans and \ai has been shown to improve collaborative performance \citep{tabrez2020mmrobotteam, hoffman-measures-for-xai}.
Especially in recent days, as model complexity increases, the explainability of predictions—through transparency, interpretability, and accessibility—has become increasingly important for enabling users to build accurate mental models that closely reflect the systems' actual capabilities~\citep{anderson2020mental, feng-boyd-graber-2022-learning, sieker-etal-2024-illusion}. %

\paragraph{Human-\mt Collaboration.}
As a widely used \ai technology, machine translation (\mt) is a common case of human-\ai collaboration in the real world, facilitating communication in multilingual settings \citep{calefato2016assessing}, and even in high-stakes domains like healthcare, law, and immigration \citep{liebling2020unmet,vieira2023ukmtusage,VieiraOHaganOSullivan2021}.
However, studies of human-\mt collaboration have historically focused on professional translators \citep{hutchins2001machine,cadwell-etal-2016-human, obrien-2024-humancentered, scansani-etal-2019-translator, sarti2025qe4pe}, and the growing population of lay users remains understudied \citep{savoldi-etal-2025-translation,kenny-etal-2022-machine}. 
Recent work in this space has focused on exploring the specificities of post-editing by non-experts \citep{koehn-2010-enabling, bawden-etal-2024-translate,obrien-etal-2018-machine},
and on developing quality feedback mechanisms to help users decide whether a given output is safe to use or not \citep{zouhar-etal-2021-backtranslation,mehandru-etal-2023-physician,ki2025isharetranslationevaluating}. However, these feedback mechanisms remain imperfect, are not available in generic translation apps, and are almost exclusively focused on text translation. This paper thus takes a complementary approach and seeks to get to the root of the problem by first measuring how people understand \mt behavior, focusing on the challenging case of speech translation. 

Furthermore, the cross-lingual \qa framework introduced here could motivate future interventions to promote \mt literacy.
While existing \mt literacy education focuses on telling users where \mt can go wrong \citep{bokerciro2019mtliteracy}, our framework lets them experience errors and their consequences for themselves in controlled settings, and calibrate their \mm through interactions.

\section{Conclusion}\label{sec:conclusion}

To understand and improve users' mental models of machine translation systems, we propose a cross-lingual \qa framework investigating how users perceive \mt error tendencies and how their mental models evolve through interaction.
Our findings show that users refine their mental models over time, with fluent and intermediate users demonstrating consistent improvements while basic users struggle to adapt. Error cues that are used to update their mental models, such as incompleteness and phonetic noise, were the most salient features, whereas topic-related errors remained the hardest to detect. Finally, transcription explanation is effective in supporting accurate mental model development, while error span highlighting encourages over-reliance.

Together, our study highlights promising directions for designing \mt systems and interfaces that better support users in building accurate and robust mental models, opening new possibilities for extending mental model research beyond \mt and toward deeper aspects of user understanding.
In particular, our experiments show that the \qa-tasked game framework is a promising interactive way to improve people's \mt literacy, especially for intermediate and bilingual users, and that existing tools can provide useful explanations in the form of transcriptions.
Yet, it opens up future challenges: how to support users with no source-language proficiency and how to encourage attention to input features that strongly influence \ai outputs.

\section*{Limitations}\label{sec:limitations}

We only experimented with French as the source language. While French offers a useful test case due to its availability in multilingual benchmarks, our findings may not fully generalize to other languages, especially those that differ more significantly from English in terms of phonetics, morphology, or domain coverage. We consider it promising that our framework shows users' ability to build mental models in French, so future work should extend this framework to a broader set of languages to study the robustness of our conclusions.

Both the number of questions and the number of participants in our experiment were limited due to resource constraints. While the small sample size can limit statistical power and the diversity of strategies observed across users, our analysis in Appendix~\ref{sec:append_add_plots} shows that the key results are statistically significant and thus meaningful. Still, scaling up the dataset and recruiting a larger, more varied participant pool would enable more reliable estimates and a richer understanding of how different user groups develop mental models.

\section*{Ethics Statement}\label{sec:ethics}
Participants were compensated \$10 per set that takes up to 40 minutes to complete, and this rate is above our region's hourly minimum wage. The institutional review board (\textsc{irb}) at our institution approved our study, confirming that potential risks were appropriately managed. Participants were explicitly informed that they would be part of a research study, and the study proceeded only after giving their consent. Recruitment was determined by participants' proficiency in English and French on the crowdsourcing platform, regardless of demographic or geographic characteristics.
We will release no \textsc{pii} from participants.
Appendix~\ref{sec:append_add_interface} has all instructions shown to participants.
We use \ai assistants to partially refine or polish writing at the sentence level (e.g., fixing grammar, re-wording sentences).

\section*{Acknowledgments}
This material is based on work supported in part by the Institute for Trustworthy AI in Law and Society (TRAILS), which is supported by the National Science Foundation under Award No. 2229885.
This material is also based upon work supported by the National Science Foundation under \abr{iis}-2403436 (Boyd-Graber) and \abr{dge}-2236417 (Balepur).
Any opinions, findings, and conclusions or recommendations expressed in this material are those of the author(s) and do not necessarily reflect the views of the National Science Foundation.
We thank the anonymous reviewers. We also thank Yu Hou, Dayeon Ki, Fenfei Guo, Hiba El Oirghi, Julio Poveda, and the members of the CLIP Lab at the University of Maryland, College Park, for their insightful feedback.

\bibliography{bib/journal-full,bib/jbg,bib/hyojung,bib/nishant}

\clearpage
\appendix

\section{Additional Details of Quality Estimation Setup and Proficiency Level }\label{sec:append_add_qe_prof}

\begin{figure}[t]
\centering
\includegraphics[width=0.48\textwidth]{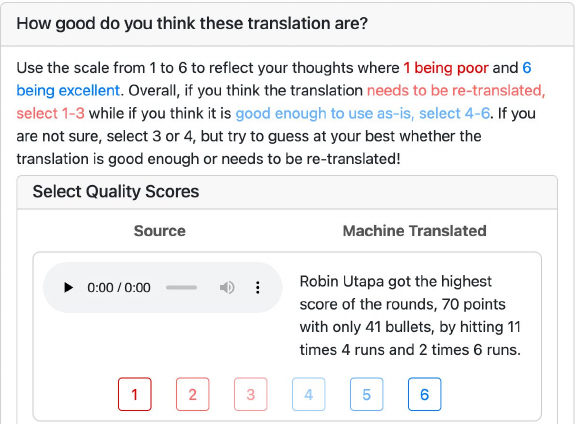} 
\caption{
Interface of the additional quality estimation task. Every four \qa items, we ask users to rate the speech translation before answering the \qa task. These \qe ratings are later used to estimate their source-language fluency.
}
\label{fig:mm4mt_interface_qe}
\end{figure}

\begin{figure}[t]
\centering
\includegraphics[width=0.48\textwidth]{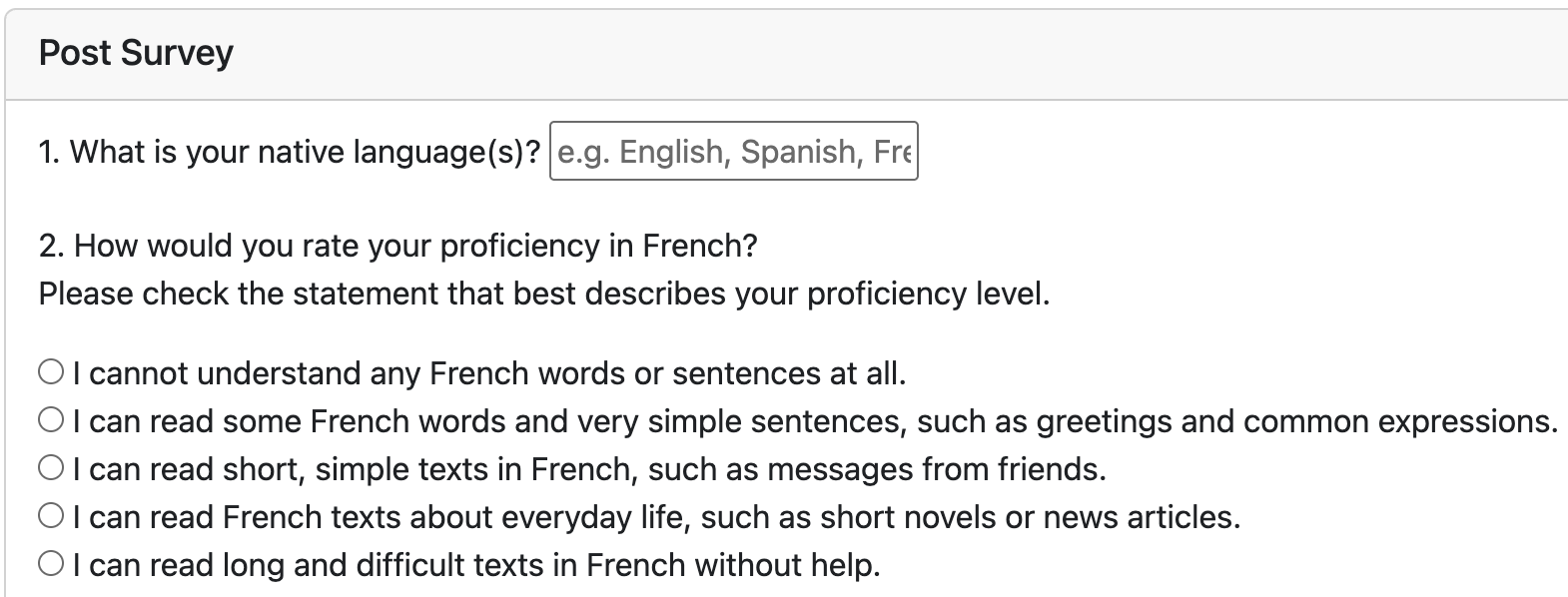} 
\caption{
Part of our post-survey, asking about French proficiency.
}
\label{fig:prof_french_survey}
\end{figure}

In our post-survey, we ask participants to self-assess their source language (French) proficiency on a five-point scale (Figure~\ref{fig:prof_french_survey}). We group the first three responses as basic, the fourth as intermediate, and the fifth as fluent.
Since self-reported language proficiency is often unreliable \citep{tomoschuk2019seven, edele2015whybothertest}, we adjust participants' proficiency levels based on their \qe performance rather than using raw survey responses. As described in Section~\ref{sec:exp_qe}, each participant completes four \qe tasks (one before every four \qa tasks), in which they rated translation adequacy on a 1–6 scale for each segment (Figure~\ref{fig:mm4mt_interface_qe}). In parallel, we compute \xcomet scores for the gold English transcription and the translated text (Section~\ref{sec:exp_interface}), converted to a 1--6 scale (Table~\ref{tab:xcomet2sc}). Participants' accumulated differences between their ratings and the \xcomet scores serve as their \qe performance.

Based on this measure, we adjust initial proficiency groups: advanced participants with accumulated differences above 30 are demoted to intermediate; intermediate participants above 40 are demoted to basic; intermediate participants below 20 are promoted to advanced; and basic participants below 30 are promoted to intermediate. The final distribution of adjusted proficiency groups appears in Table~\ref{tab:participants_dist}.

\begin{table}[h]
\centering
\begin{tabular}{ll}
\toprule
Range of $x$ & Scale \\
\midrule
$x < 0.60$ & 1 \\
$0.60 \leq x < 0.80$ & 2 \\
$0.80 \leq x < 0.90$ & 3 \\
$0.90 \leq x < 0.95$ & 4 \\
$0.95 \leq x < 0.98$ & 5 \\
$x \geq 0.98$ & 6 \\
\bottomrule
\end{tabular}
\caption{Mapping from \xcomet score $x$ to 1--6 scale.}
\label{tab:xcomet2sc}
\end{table}

\section{Statistical Significance}\label{sec:append_add_plots}
\begin{figure}[t]
    \centering
    \includegraphics[width=0.4\textwidth]{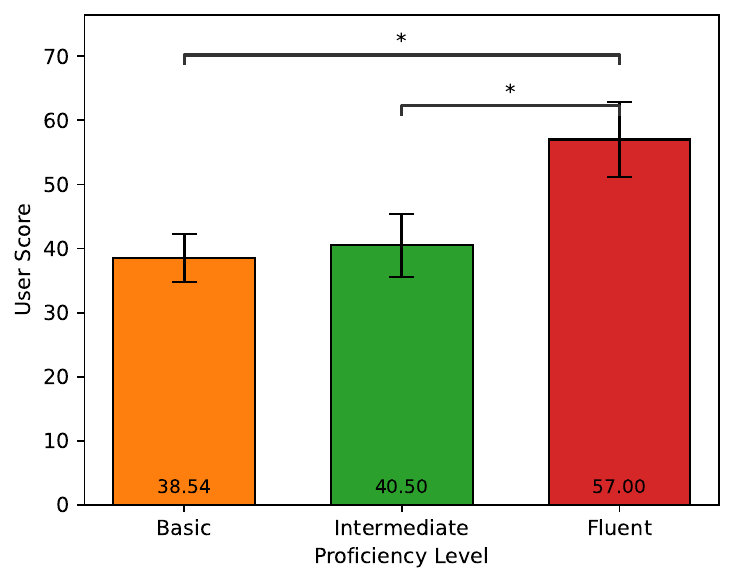} 
    \caption{
    User scores across proficiency levels. * indicates p-value < $0.05$.
    }
    \label{fig:bar_prof_user_score}
\end{figure}

\begin{figure}[t]
    \centering
    \includegraphics[width=0.4\textwidth]{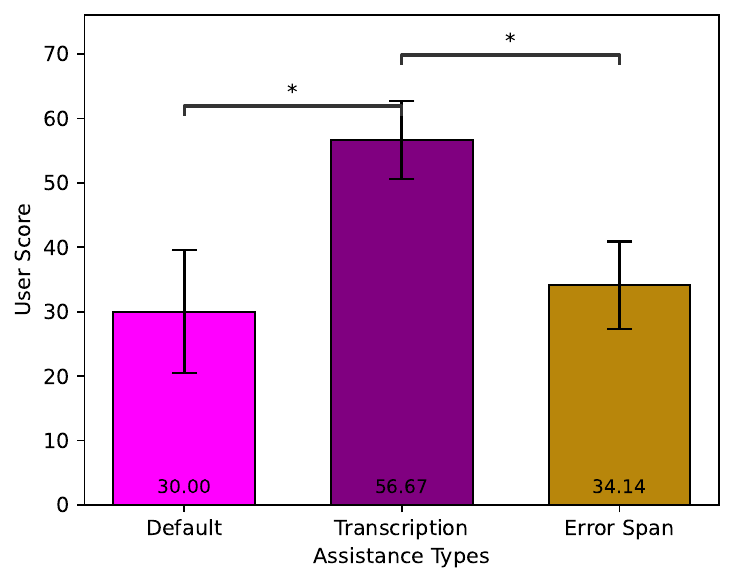} 
    \caption{
    User scores across explanation conditions for intermediate participants. * indicates p-value < $0.05$.
    }
    \label{fig:bar_user_score}
\end{figure}

We present statistical tests of user scores across proficiency levels (Figure~\ref{fig:bar_prof_user_score}). Scores for fluent users differ meaningfully from those of both basic and intermediate groups, while intermediate users perform slightly better than basic users, though the difference is not substantial.
To examine the role of explanation, we compare user scores across assistance conditions (Figure~\ref{fig:bar_user_score}), focusing on the intermediate group to control for proficiency effects. Transcription explanation shows clear improvement over the default condition and performs better than error span explanations, whereas error span yields higher scores than the no-explanation baseline, but the difference is not statistically meaningful. This supports our conclusion that transcription explanation effectively helps users refine their mental models, whereas error spans are less suitable for fostering well-formed ones.


\section{Details of Interface}\label{sec:append_add_interface}
We provide the main instructions shown to the users (\ref{prompt:main_instruction}) and the instructions for collecting the reflection notes (\ref{prompt:reflection_note}).

\section{Dataset Details}

For our studies, we used 16 reading comprehension sets from
\textsc{2m-belebele} \cite{2mbelebele}.
All data is in English or French.
The dataset is publicly available and our usage in this paper is thus within its intended use.
We did not collect the dataset ourselves, so we did not check it for \textsc{pii}.

\begin{prompt}[title={\thetcbcounter: Main Instruction}, label=prompt:main_instruction]
Machine translated passage will be consumed by the question answering (\qa) system to answer the question. Select only the sentences that you believe need to be re-translated by a professional translator because they may have \mt errors in key information, so that the \qa system can answer the question correctly with the final passage. If you think a machine translated sentence is good enough to use as-is, it's better not to select it to maximize your reward: even in some cases, selecting nothing is okay.
\end{prompt}

\begin{prompt}[title={\thetcbcounter: Instruction for Reflection Note}, label=prompt:reflection_note]
Playing the game with these questions in mind will help you make better decisions. We encourage you to jot down a brief note after each item as a reflection. You'll revisit these same questions at the end of the game.

1. What kinds of things do you think the \mt system used here is likely to handle correctly or incorrectly?

2. What types of errors do you think this \mt system is prone to make?

3. What characteristics have you learned to look for when identifying incorrect translations?
\end{prompt}

\end{document}